\documentclass[conference]{IEEEtran}
\usepackage{cite}
\usepackage{graphicx,epstopdf,url}
\usepackage{subfigure}
\usepackage{verbatim}
\usepackage{textcomp}
\usepackage{slashbox}
\usepackage{amsmath}
\usepackage{amssymb}
\usepackage{multirow}

\usepackage{xcolor}
\usepackage{soul}
\usepackage{tabularx}
\usepackage{makecell}
\usepackage{wrapfig}

\usepackage{booktabs}
\usepackage[utf8]{inputenc}  
\usepackage{indentfirst}
\usepackage[all]{xy}
\usepackage{bbm}
\usepackage{algorithm}
\usepackage{algpseudocode}

\usepackage{graphicx}

\newcommand{\our}{I-SplitEE}

\def\BibTeX{{\rm B\kern-.05em{\sc i\kern-.025em b}\kern-.08em
    T\kern-.1667em\lower.7ex\hbox{E}\kern-.125emX}}
\begin{document}
\title{I-SplitEE: Image classification in Split Computing DNNs with Early Exits}

\author{
    \IEEEauthorblockN{Divya J. Bajpai, Aastha Jaiswal, and
    Manjesh K. Hanawal\\
   \IEEEauthorblockA{
    Indian Institute of Technology Bombay, India}
    Email: \{20i190003, 22n0463, mhanawal\}@iitb.ac.in}
}
\maketitle


\begin{abstract}
The recent advances in Deep Neural Networks (DNNs) stem from their exceptional performance across various domains. However, deploying these networks on resource-constrained devices—like edge, mobile, and IoT platforms—is hindered by their inherent large size. Strategies have emerged, from partial cloud computation offloading (split computing) to integrating early exits within DNN layers. Our work presents an innovative unified approach merging early exits and split computing. We determine the 'splitting layer,' the optimal depth in the DNN for edge device computations, and whether to infer on edge device or be offloaded to the cloud for inference considering accuracy, computational efficiency, and communication costs.
Also, Image classification faces diverse environmental distortions, influenced by factors like time of day, lighting, and weather. To adapt to these distortions, we introduce I-SplitEE, an online unsupervised algorithm ideal for scenarios lacking ground truths and with sequential data. Experimental validation using Caltech-256 and Cifar-10 datasets subjected to varied distortions showcases I-SplitEE's ability to reduce costs by a minimum of $55\%$ with marginal performance degradation of at most $5\%$.
\end{abstract}

\section{Introduction}
In recent years, the scale of Deep Neural Networks (DNNs) has undergone substantial expansion, leading to remarkable performance achievements \cite{han2022survey}, especially for computer vision tasks such as image classification \cite{bochie2021survey}. This increased scale has increased computational demands, rendering their deployment on resource-constrained platforms like mobile and edge devices. To mitigate the computational challenges associated with deploying DNNs on these devices, various strategies have been introduced. These include approaches like Split Computing, Early Exits, and offloading to cloud \cite{matsubara2022split}.\\
\textbf{Split Computing. } Edge devices, often constrained by limited processing capabilities, can opt to offload their data to the cloud, which boasts superior computational power. This approach involves running the full-fledged Deep Neural Network (DNN) in the cloud, subsequently returning the inferences to the edge device. However, this offloading of data to the cloud incurs latency due to communication delays between the edge device and the cloud. To harness the combined resources available on both mobile devices and the cloud, a strategy known as "split computing" has been introduced. In split computing, the DNN is bifurcated into two segments: a subset of initial layers is deployed on the edge device, while the remaining layers are hosted in the cloud. The portion deployed on the edge device serves as an encoder for processing input images, reducing their size and, consequently, mitigating latency costs. This hybrid approach facilitates edge-cloud co-inference. Nonetheless, it's important to note that in split computing, all samples still need to be inferred at the final layer on the cloud, which, while reducing the computational burden on edge devices, maintains significant communication costs, as every sample is offloaded \cite{matsubara2022split}.\\
\textbf{Early Exits (EE-DNNs).} Several previous methods, including BranchyNet \cite{teerapittayanon2016branchynet}, SPINN \cite{laskaridis2020spinn}, and LGViT \cite{xu2023lgvit}, enable the classification of input images by adding classifiers at intermediate layers within DNNs. In these early exit DNNs, the decision to classify the image at an intermediate stage is contingent on the confidence of the prediction surpassing a predefined threshold. Exiting the network at its early layers demands less computational resources but may lead to less precise predictions. Conversely, deeper exits offer greater precision but at the cost of higher computational demands. Therefore, determining the best exit point within the network is important to model the accuracy-efficiency trade-off.

In our approach, we introduce a hybrid strategy that combines both split computing and early exits to optimize resource usage on both edge devices and the cloud. We define the "splitting layer" as the final layer on the edge device. At this layer, we attach an exit point for making inferences locally on the edge device. Before an input image is transferred to the cloud, it undergoes classification on the edge device. If the prediction confidence exceeds a predetermined threshold, the process is terminated, and the image is classified on the edge. In cases where the confidence falls below the threshold, the image is offloaded to the cloud for further processing.

While previous methods \cite{pacheco2021early, shao2020bottlenet++} traditionally employ a fixed splitting layer on the edge device, our approach advocates for the adaptive selection of the splitting layer. This adaptability is particularly crucial in real-world scenarios where input images often exhibit various distortions \cite{dodge2016understanding}. For instance, edge devices frequently encounter noisy or blurred images, with different levels of distortion. Pacheco et al. \cite{pacheco2021distorted} empirically illustrate the influence of noise on confidence values in images. The distortion in images changes the latent distribution of the dataset and significantly impacts the confidence values and inference results \cite{pacheco2021early}. During inference, the data arrives in an online (sequential order) fashion and the ground truth values for the image sample are not available. Hence the question then arises, how to adaptively choose the optimal splitting layer to model the accuracy-efficiency trade-off when the underlying latent distribution of incoming data changes?

To address the challenge of selecting an optimal splitting layer when the test data has a different distribution as compared to the dataset used for training, we introduce an online learning algorithm called \our{}. This algorithm leverages the Multi-Armed Bandit framework \cite{auer2002finite}, utilizing the confidence in prediction as a surrogate for prediction accuracy at the exit point. The computational cost is the cost of running the DNN on edge devices, while the offloading cost is the cost of communication delays between edge and cloud. The primary objective is to maximize prediction confidence while minimizing computational and offloading costs. \our{} dynamically determines the optimal splitting layer for an incoming sample based on past observations from the dataset with a given latent data distribution. \our{} model could be adapted to different edge devices and communication networks as well using user-specific inputs.

For our backbone model, we adopt the lightweight MobileNet V2 \cite{sandler2018mobilenetv2} model. This choice is an ideal testbed for our algorithm, given its efficiency and competitive accuracy compared to state-of-the-art models. We conduct experiments under varying levels of noise (distortion) in the dataset, as detailed in Section \ref{sec: experiments}. Our experimental results on Caltech-256 \cite{griffin2007caltech} and Cifar-10 \cite{krizhevsky2014cifar} datasets demonstrate that the optimal splitting layer differs as there are changes in the distribution of data. Specifically, \our{} achieves a significant reduction in the cost ($>55\%$) with only a minimal drop in performance ($\leq 5\%$) when compared to the scenario where all samples are processed at the final layer.

Our key contributions encompass: 1) We integrate split computing and early exiting to enable early inferences for "easy" samples on edge devices. 2) We develop an online learning algorithm \our{} to dynamically select the optimal splitting layer based on context in an online and unsupervised manner. 3) \our{} optimizes the resource utilization across edge devices and the cloud. 4) We experimentally validate that \our{} minimizes performance degradation while significantly reducing costs compared to the final exit scenario.

\begin{figure}
    \centering
    \includegraphics[scale = 0.5]{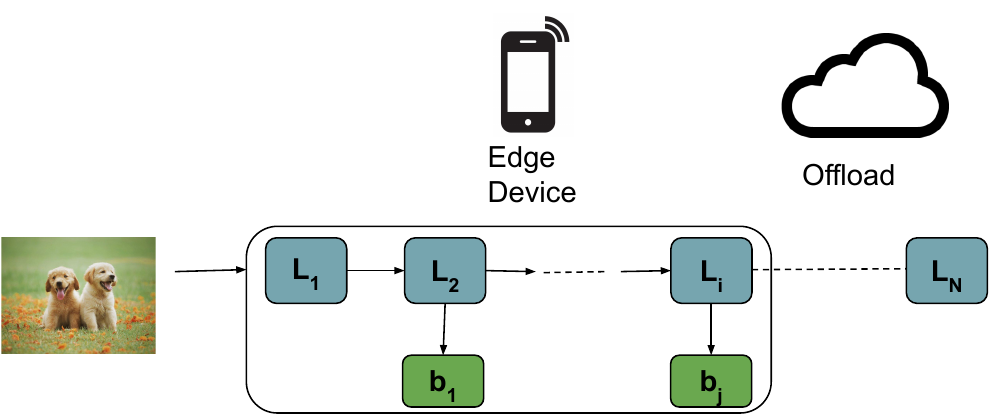}
    \caption{Split computing combined with early exits, where the DNN is split into two parts with an additional advantage of inference on mobile device.}
    \label{fig:enter-label}
\end{figure}

\label{sec:intro}

\section{Related works}
In this section, we discuss the previous works on Split computing and Early Exit to use DNNs on mobile devices. 

\textbf{Split computing for DNNs.}
Neurosurgeon, as introduced in \cite{kang2017neurosurgeon}, explores strategies for optimizing the splitting of DNNs based on the cost considerations associated with selecting a specific splitting layer. In a similar vein, BottleNet \cite{eshratifar2019bottlenet} incorporates a bottleneck mechanism within split computing. This approach entails deploying a segment of the DNN on an edge device to encode the input sample into a more compact representation before transmitting it to the cloud for further processing.
Various training methodologies have been proposed for training the encoder situated on the edge device. Notably, BottleNet++ employs cross-entropy-based training approaches in the context of split computing.\\
\textbf{Early-Exit DNNs.} 
Early-exit DNNs have found applications across diverse tasks. In the context of image classification, BranchyNet \cite{teerapittayanon2016branchynet} and several preceding studies utilize classification entropy metrics at different intermediate layers to determine whether early inference can be made with sufficient confidence. Approaches like SPINN \cite{laskaridis2020spinn} incorporate early exits into DNN architectures, primarily aimed at handling service disruptions. Beyond image classification, early exits have found relevance in various domains, spanning natural language processing tasks (NLP) \cite{xin2020deebert, liu2021elasticbert}.\\
\textbf{DNNs on mobile device.} AdaEE \cite{pacheco2021early} employs a combination of early-exit DNNs and DNN partitioning to facilitate offloading data from mobile devices using early-exit DNNs. 
LEE (Learning Early Exit) \cite{ju2021learning}, DEE (Dynamic Early exit) \cite{ju2021dynamic} and UEE-UCB (Unsupervised early exits using Upper Confidence Bound) \cite{hanawal2022unsupervised} leverage the multi-armed bandit framework to determine optimal exit points. However, they lack the offloading and primarily focus on performing operations on edge devices. LEE and DEE are specifically designed for efficient edge device inference, particularly in cases of service disruptions, employing utility functions that require access to ground-truth labels. Notably, UEE-UCB specializes in learning the optimal exit strategy for the ElasticBERT \cite{liu2021elasticbert} backbone, primarily tailored for text classification tasks.

Our approach distinguishes itself from prior methods in several key aspects. 1) We consider not only accuracy but also computational and communication costs when determining the optimal splitting layer. 2) Our method operates in an online and unsupervised setup, offering a novel perspective. 3) We leverage contextual information, such as confidence scores, to dynamically determine the splitting layer for each sample during edge device processing.

\section{Problem Setup}\label{sec: setup}
In our scenario, we train a deep neural network (DNN) consisting of $L$ layers. We introduce exit points into this backbone after specific layers. We denote the set of layers as $[L] = \{1, 2, \ldots, L\}$. The set of target classes for classification is denoted as $\mathcal{C}$. For a given image, $x$, and an exit attached to layer $i$, we define $\hat{P}_i(c)$ as the estimated probability that $x$ belongs to class $c$, where $c$ belongs to $\mathcal{C}$. The confidence in the estimated probability class, defined as $C_i = \max_{c \in \mathcal{C}}\hat{P}_i(c)$, is the maximum among these estimated probabilities.

The DNN processes the image $x$, and the point at which the DNN splits can occur at any layer $i$ where exit is attached, where the layers $1$ through $i$ reside on the edge device, and the remaining layers $i+1$ through $L$ are processed on the cloud. In our setup, each image requires two-stage decisions: 1) Determining the optimal splitting point in the DNN, and 2) Deciding whether to exit at the chosen splitting layer or offload the task. Notably, the choice of the splitting layer relies on the collective context rather than individual samples. The choice to exit or offload depends on individual samples and follows this process: When the DNN splits at the $i$th layer, we calculate $C_i(x)$ and compare it with a predefined threshold, denoted as $\alpha$. If the confidence $C_i(x)\geq\alpha$, the image sample is inferred locally on the edge device and does not require offloading. Otherwise, it is offloaded to the cloud for inference at the final layer incurring communication costs. 

The cost of using the DNN up to layer $i$ encompasses the computational cost of processing the image sample up to that layer on the edge and performing inference. We represent the computational cost associated with splitting at the $i$th layer as $\gamma_i$, where $\gamma_i$ is proportional to the layer depth $i$. Additionally, we introduce the cost of offloading from the edge device to the cloud, denoted as $o$, which depends on factors like transmission type (\textit{e.g.}, 3G, 4G, 5G, or Wi-Fi). We define the reward function for splitting at layer $i\in [L]$ as 

\begin{equation}
\label{eq: Reward1}
    r(i) = \left\{
        \begin{array}{ll}
            C_i-\mu\gamma_i & \textit{if} \quad C_{i} \geq \alpha \text{ or } i=L\\
             C_L- \mu(\gamma_i+ o) & \textit{otherwise,}
        \end{array}
    \right.
\end{equation}
where $\mu$ serves as a conversion factor, enabling us to represent the cost in terms of confidence. The user-defined value of $\mu$ reflects the user's preference and captures the balance between accuracy and cost.

The reward function's interpretation is as follows: when the DNN generates a confident prediction at the splitting layer, the reward is calculated as the confidence at the splitting layer minus the cost associated with processing the sample up to the $i$th layer and performing inference. In the event of lower confidence, the sample is offloaded to the cloud, incurring offloading costs and inferring at the final layer, with the confidence of $C_L$. If $i$ is equal to $L$, the sample is processed entirely on the edge device, eliminating the need for offloading. For any splitting layer $i \in [L]$, the expected reward can be expressed as
\begin{multline}
\mathbb{E}[r(i)] = \mathbb{E}[C_i-\mu\gamma_i|C_i\geq \alpha]\cdot P[C_i\geq\alpha]\\ + \mathbb{E}[C_L-\mu(\gamma_i+o)|C_i<\alpha]\cdot P[C_i<\alpha]
\end{multline}
and for the last layer $L$, it is a constant given as $\mathbb{E}(r(L)) = C_L-\mu\gamma_L$. The goal is to find the optimal splitting layer $i^{*}$ defined as  $i^{*} = \arg \max_{i\in [L]}\mathbb{E}[r(i)]$.

Recall (section \ref{sec:intro}) that the problem of choosing optimal thresholds is in an online and unsupervised setup. In the context of online learning problems, the Multi-Armed Bandit (MAB) \cite{auer2002finite} setup plays a pivotal role in addressing challenges related to dynamic decision-making. Hence, we employ an MAB framework to find the optimal splitting layer based on user-specific arguments. In this framework, we establish our action set as the set of layer indices where exits are attached in the DNN denoted as $E$, where each choice corresponds to splitting the DNN at an exit point. In MAB terminology, these choices are referred to as "arms." Within this framework, we devise a policy denoted as $\pi$ that selects arm $i_t$ at each time step $t$ based on prior observations. We define the cumulative regret of policy $\pi$ over $T$ rounds as
\begin{equation}
    R(\pi, T) = \sum_{t = 1}^{T}\mathbb{E}[r(i^{*}) - r(i_t)]
\end{equation}
where the expectation is with respect to the randomness in the arm selection caused by previous samples. A policy $\pi^{*}$ is said to be sub-linear if the average cumulative regret vanishes, i.e. $R(\pi^{*}, T)/T \rightarrow 0$.

\section{Algorithm}
In this section, we develop an algorithm named \our{}. The algorithm is based on the 'optimism in the face of uncertainty principle' and utilizes the upper confidence bounds.

\begin{algorithm}[H]
        \caption{\our{}}
        \begin{algorithmic}[1]
          \State\textbf{Input:} $\alpha \text{ (threshold)}, \beta\geq 1, E, \gamma_i \text{ } \forall i\in E$, $o$
\State \textbf{Initialize:} $Q(i)\gets 0, N(i)\gets 0$.
\State Initialize by playing each arm once.
\For{$t = |E|+1, |E|+2, \ldots$}
\State Observe an instance $x_t$ 
\State $i_t \gets \arg \max_{i\in E}\left(Q(i)+\beta\sqrt{\frac{\ln(t)}{N(i)}}\right)$
\State Pass $x_t$ till layer $i_t$, use threshold $\alpha$ and observe $C_{i_t}$

\If{$C_{i_t}\geq \alpha$} 
    \State  Infer at layer $i_t$ and exit
    \State $r_t(i_t) \gets C_{i_t}(x_t)-\mu\gamma_{i_t}$, $N_{t}(i_t) \leftarrow N_{t-1}(i_t)+1$
    \State $Q_{t}(i_t) \leftarrow \sum_{j=1}^{t}r_{j}(k)\mathbbm{1}_{\{k=i_t\}}/N_{t}(i_t)$ 
\Else 
    \State Offload to the last layer. Observe $C_L$
    \State $r_{t}(i_t) \gets C_{L}(x_t)-\mu(\gamma_{i_t} + o)$
    \State $N_{t}(i_t) \leftarrow N_{t-1}(i_t)+1$
    \State $Q_{t}(i_t) \leftarrow \sum_{j=1}^{t}r_{j}(k)\mathbbm{1}_{\{k=i_t\}}/N_{t}(i_t)$  
    \State Infer at the last layer
\EndIf
\EndFor
\end{algorithmic}
\label{alg:algorithm}
\end{algorithm}
\begin{figure*}
    \centering
    \includegraphics[scale = 0.35]{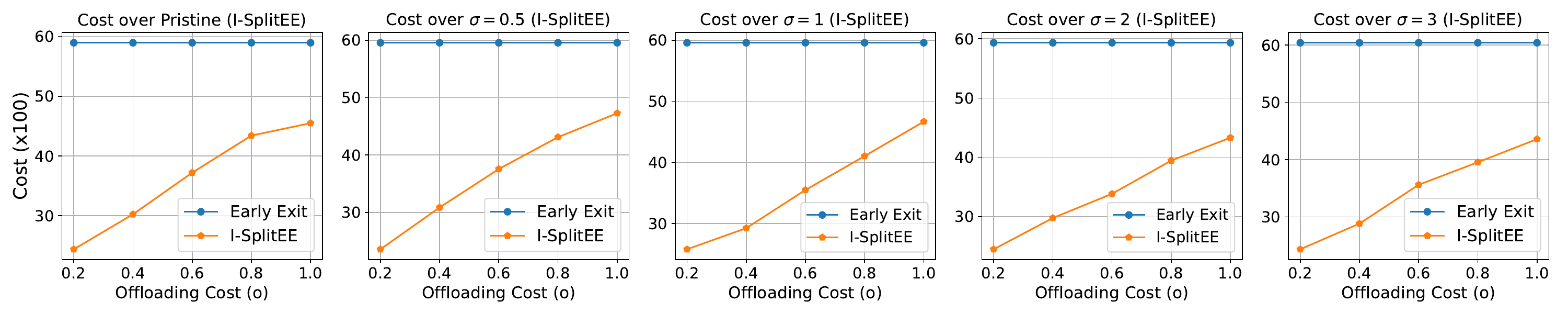}
    \caption{Effect of different offloading costs over the total cost. The cost is mostly linear but with some deviations.}
    \label{fig:cost}
\end{figure*}

\begin{figure*}
    \centering
    \includegraphics[scale = 0.35]{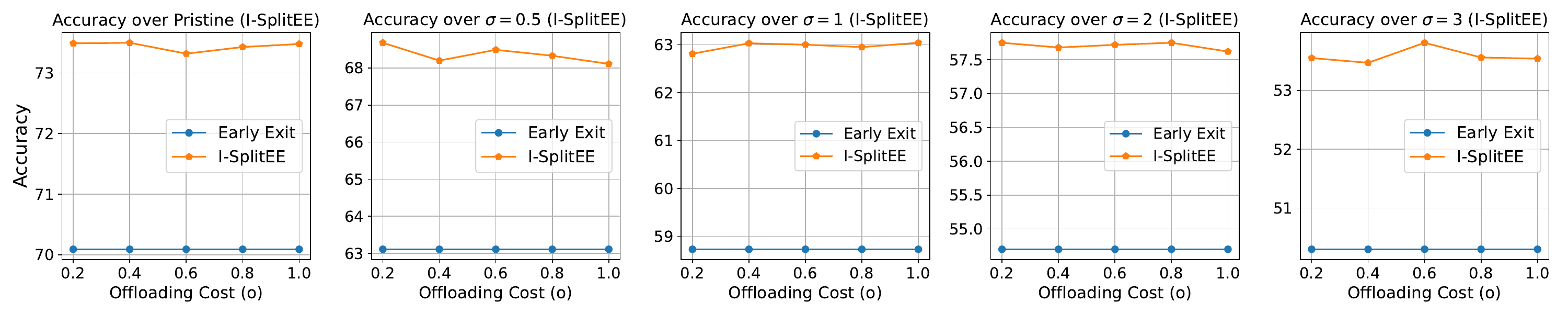}
    \caption{Effect of different offloading costs over the accuracy. The accuracy remains constant, however, the changes might be due to randomness in the dataset.}
    \label{fig:accuracy}
\end{figure*}
The input to the algorithm is the exploration parameter $\beta$, confidence threshold $\alpha$, number of exits $E$, computational cost $\gamma_i \text{ }\forall i \in E$, and the offloading cost $o$. The pseudo-code of the algorithm is given in Algorithm \ref{alg:algorithm}. The working of the algorithm could be explained as follows: It plays each arm once to obtain rewards $Q(i)$ and counters $N(i)$ for each layer with an attached exit. After observing each arm once, it plays an arm $i_t$ that maximizes the UCB index (line 6) in succeeding rounds. UCB index comprises the weighted sum of empirical averages of rewards $Q(i)$ and the confidence bonuses. The sample is then processed till the splitting layer $i_t$ and if the confidence is above the threshold $\alpha$, the sample exits the DNN; else the sample is offloaded to the cloud with additional cost $o$. 

From the analysis of UCB1 \cite{ML02_UCB1_Auer}, one can easily verify that \our{} achieves a sub-linear regret. The observed regret is $\mathcal{O}\left(\sum_{i \in \mathcal{A}\backslash i^{*}}\frac{log(n)}{\Delta_{i}}\right)$ where $\Delta_i = r(i^*)-r(i)$.

\section{Experiments}\label{sec: experiments}

\begin{table*}[]
\centering
\caption{In this table, we provide the $\%$ reduction in cost as well as accuracy with different distortion levels $\sigma$ on Caltech-256 dataset.}
\begin{tabular}{ccccccccccccc}
\hline
\textbf{Model/Distortion} & \multicolumn{2}{c}{\textbf{$\sigma = 0.0$}} & \multicolumn{2}{c}{\textbf{$\sigma = 0.5$}} & \multicolumn{2}{c}{\textbf{$\sigma = 1$}} & \multicolumn{2}{c}{\textbf{$\sigma = 1.5$}} & \multicolumn{2}{c}{\textbf{$\sigma = 2$}} & \multicolumn{2}{c}{\textbf{$\sigma = 3$}} \\ \hline
                       & \textbf{Acc}       & \textbf{Cost}       & \textbf{Acc}       & \textbf{Cost}       & \textbf{Acc}      & \textbf{Cost}      & \textbf{Acc}       & \textbf{Cost}       & \textbf{Acc}      & \textbf{Cost}      & \textbf{Acc}      & \textbf{Cost}      \\ \hline
\textbf{Final layer}   & 78.5              & 100                & 75.2               & 100                & 71.6             & 100               & 69.8               & 100                & 67.8             & 100               & 64.1             & 100               \\
\textbf{Random exit}   & -4.1               & -33.5\%               & -3.4               & -20.1\%               & -5.1              & -39.7\%              & -7.2               & -41.3\%               & -6.2              & -37.8\%              & -9.5              & -43.9\%              \\
\textbf{AdaEE}         & -3.3               & -45.8\%               & -4.2               & -47.1\%               & -4.6              & -47.4\%              & -5.8               & -49.7\%               & -5.7              & -53.0\%                & -5.2              & -50.2\%              \\
\textbf{Early-Exit}    & -5.3               & -39.2\%               & -9.1               & -38.5\%               & -9.9              & -38.5\%              & -10.5              & -38.8\%               & -10.1             & -38.4\%              & -10.8             & -37.8\%              \\ \hline
\textbf{I-SplitEE}       & \textbf{-2.0}        & \textbf{-66.3\%}      & \textbf{-2.1}      & \textbf{-65.9\%}      & \textbf{-3.9}     & \textbf{-66.7\%}     & \textbf{-5.0}        & \textbf{-67.4\%}      & \textbf{-4.8}     & \textbf{-66.4\%}     & \textbf{-4.7}     & \textbf{-65.7\%}     \\ \hline
\end{tabular}
\label{tab: results}
\end{table*}

In this section, we provide all the experimental details of the paper. There are two key phases in our experimental setup.
The source code is available at \url{https://anonymous.4open.science/r/I-SplitEE-5DE5/README.md}.\\
\textbf{1) Training the backbone: } To evaluate \our{}, we train the MobileNet V2 model with $6$ task-specific exits at intermediate points and the final exit. The placement of exits is strategically chosen as described in some previous works \cite{wang2019dynexit, pacheco2021early}. The dataset used for training is the Caltech-256 \cite{griffin2007caltech} and Cifar-10 \cite{krizhevsky2014cifar} datasets. We closely follow the training procedure as described in the paper \cite{pacheco2021early, ju2021learning}. The Caltech-256 dataset contains undistorted images of objects such as bikes, bears, camels etc. The dataset has a total of 257 different classes. The Cifar-10 dataset consists of 10 classes and also has a classification task. For the training of the backbone, we split the dataset into $80\%$ for training, $10\%$ for validation and remaining as the test set. Note that after this step no training happens. Except in table \ref{tab: cifar data}, all the given results are from the Caltech-256 dataset due to space constraints.\\
\textbf{2) Unsupervised learning of Splitting layer based on underlying distribution:} The change in distribution can be defined by multiple environmental effects, however, we focus on the change as image distortion. Following the training of the DNN with attached exits, we apply Gaussian noise as image distortion on each image from the test set. Gaussian noise occurs when the camera goes out of focus or the image is from a foggy environment. To make an image noisy, we add Gaussian noise to an image with zero mean and standard deviation $\sigma$ to an undistorted image. More standard deviation adds more noise to an image. We vary the standard deviation as $\sigma \in \{0.5, 1, 1.5, 2, 2.5, 3\}$. 

After adding the noise to the image we employ \our{} to find the optimal splitting layer based on the change in distortion present in the image. The predefined threshold is learned on the validation set of the model and is set to $0.6$ for Caltech and $0.9$ for Cifar-10 as learned by the model. We also have two types of costs: 1) Computational cost: As explained in section \ref{sec: setup}, the computational cost is directly proportional to the number of layers being processed on the edge device i.e. $\gamma_i = \lambda i$ where $\lambda$ is the computational cost for a single layer. $\lambda$ is user-defined and depends on the processing power of the user's device. We experimented with different processing cost values but due to space constraints, the results are not provided. However, we set $\lambda = 0.1$ thus normalizing the cost for direct comparison. 2) Offloading cost is also user-defined and depends on the communication network used (\textit{e.g.} 3G, 4G, 5G and Wi-Fi). Hence, we experiment with different offloading costs from the set $o\in \{0.2, 0.4, \ldots, 1.0\}$ and show (see subsection \ref{sec:offloading}) that \our{} gets better results from other baselines making it robust to different communication networks. For the main results in table \ref{tab: results}, we fix the offloading cost as $o = 1.0$ which is the worst-case scenario. More details on how to compute the offloading cost can be found in \cite{kuang2019partial}.

We run all the experiments $10$ times and average the results. The action set is the set of attached exits $\mathcal{A} = \{3, 6, \ldots 18, 20\}$. We set the trade-off factor $\mu = 1$ which could be varied based on the user's preferences for accuracy and cost. Smaller values of $\mu$ will provide accurate results but with higher costs and vice versa. All the experiments are performed on a single NVIDIA RTX 2070 GPU. The training requires $\sim10$ hours of GPU runtime for Caltech-256 and $\sim7$ hours for Cifar-10 with all the exits. Learning of the optimal splitting layer could be executed quickly using the CPU. 
\subsection{Baselines}
\textbf{1) Final-layer:} We form this as our main baseline as it is similar to the MobileNet's inference. However, added exits might marginally deteriorate the performance.\\
\textbf{2) Random exit:} We randomly choose a splitting layer and then exploit that layer for the complete dataset. We choose the random splitting layer for $10$ times and then average the results provided in table \ref{tab: results}.\\
\textbf{3) AdaEE: }  fixes the splitting layer and then adaptively chooses the threshold as given in the paper \cite{pacheco2021early}. There is an option to offload but the splitting layer is fixed. All the hyperparameters were kept the same as in the paper.\\
\textbf{4) Early-Exit:} is similar to the classic early exiting setup where all the samples are inferred at the edge device and based on the confidence values being above a given threshold. The sample passes through the backbone until it obtains sufficient confidence.

\subsection{Impact of change in distortion} To gain a deeper understanding of the influence of image distortion, we conducted an analysis of the confidence values associated with the final layer of our model. 
\begin{wrapfigure}[15]{r}{0.31\textwidth}
  \begin{center}
    \includegraphics[width=0.29\textwidth]{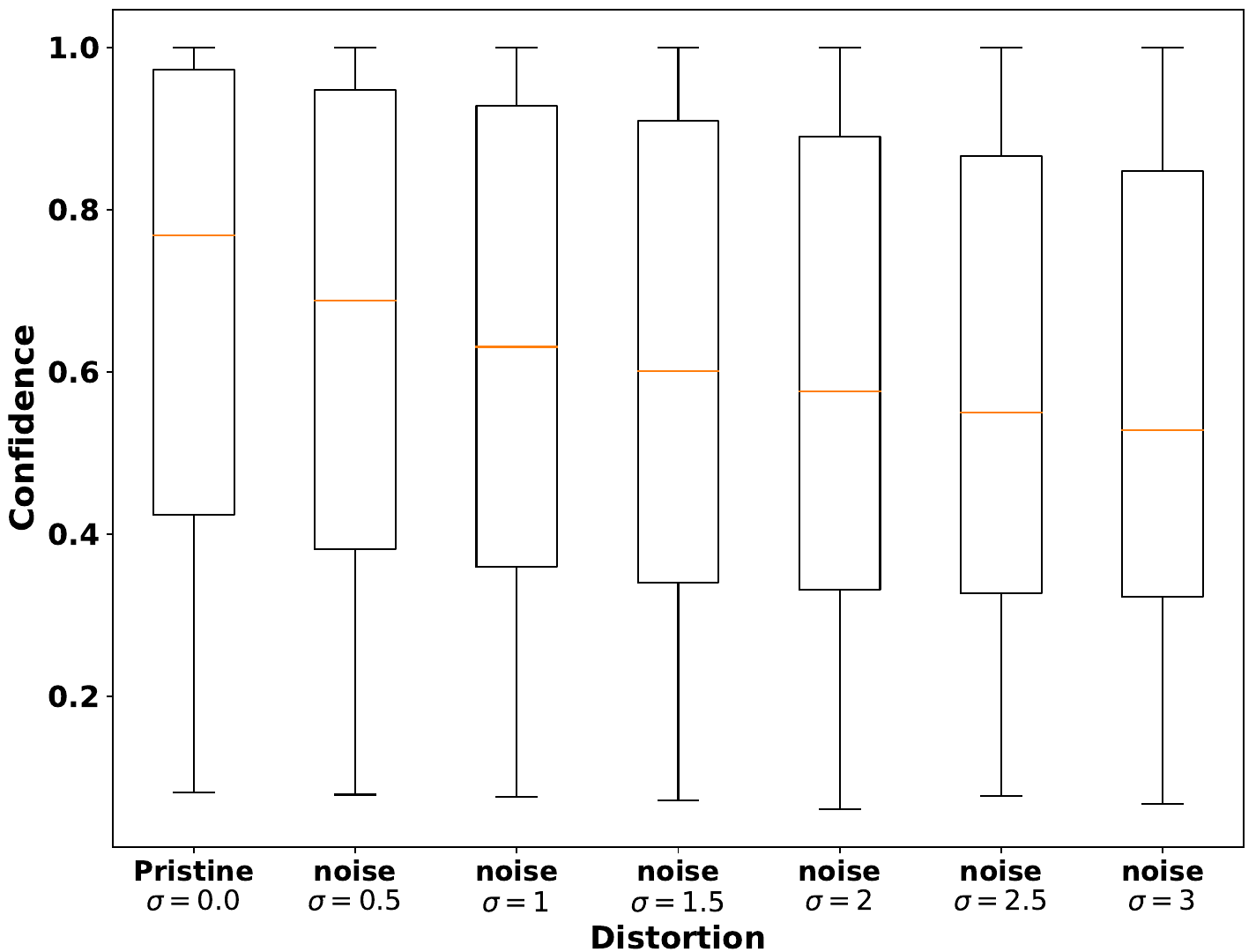}
  \end{center}
  \caption{Effect of added noise on confidence. These confidence values are the final layer.}
  \label{fig: confidence}
\end{wrapfigure}
As depicted in Figure \ref{fig: confidence}, we present boxplots that showcase the confidence values at the final layer, considering various levels of image distortion, as well as undistorted images. The visual representation in this figure serves as a compelling illustration of the significant impact that image distortion exerts on the model's confidence. There is a loss in confidence as the distortion in images increases and might affect the model if a fixed splitting layer is used.
This observation compels us to adaptively choose the splitting layer as with a change in the underlying distribution of the dataset the distribution of confidence values also changes at different exits including the final exit.

\subsection{Analysis over different overheads}\label{sec:offloading}
Given that the offloading cost is a user-defined parameter, we conducted a comprehensive analysis by varying this cost within the set of values $\{0.2, 0.4, 0.6, 0.8, 1.0\}$ in figure \ref{fig:cost}. This exploration focused on examining the behaviour of both accuracy and cost under different offloading cost scenarios.
As the offloading cost increases, it is unsurprising that the overall cost also rises. Notably, this increase in the offloading cost does not result in a significant impact on accuracy. This observation suggests the robustness of \our{} to variations in offloading cost, ensuring that accuracy remains relatively stable.
However, we observe that the choice of the optimal splitting layer adjusts with shifts in the offloading cost. When the offloading cost is elevated, \our{} opts for a deeper splitting layer, aligning with expectations, as the computational cost becomes relatively smaller compared to the offloading cost. Consequently, the algorithm's design prompts the sample to be locally inferred on the edge device. Remarkably, this decision does not compromise the accuracy of the model, due to offloading of samples with low confidence on the edge device. As a result, every prediction maintains a high level of confidence, resulting in only a marginal reduction in accuracy.

Moreover, our analysis reveals that even from a cost perspective, our model outperforms the early-exit baseline. This finding strengthens the claims regarding the efficacy of \our{} and its ability to strike a compelling balance between accuracy and cost.

\section{Results}
\begin{table}[]
\caption{Similar results as in Table \ref{tab: results} for Cifar-10. Observe that reduction in accuracy and cost is in $\%$}
\begin{tabular}{ccccccc}
\hline
\textbf{Model/Distortion} & \multicolumn{2}{c}{\textbf{$\sigma$ = 0.0}} & \multicolumn{2}{c}{\textbf{$\sigma$ = 0.5}} & \multicolumn{2}{c}{\textbf{$\sigma$ = 1}} \\ \hline
\textbf{}                 & \textbf{Acc}        & \textbf{Cost}      & \textbf{Acc}       & \textbf{Cost}       & \textbf{Acc}      & \textbf{Cost}      \\ \hline
\textbf{Final layer}      & 92.5                & 100                & 89.9               & 100                 & 78.3              & 100                \\
\textbf{Random exit}      & -0.9                & -37.2              & -9.1               & -41.8              & -8.7              & -31.5              \\
\textbf{AdaEE}            & -0.3                & -65.9              & -4.5               & -67.2               & -6.3              & -41.5              \\
\textbf{Early-Exit}       & -0.7                & -64.4              & -8.9               & -65.6               & -9.5              & -28.7              \\ \hline
\textbf{\our{}}          & \textbf{-0.05}      & \textbf{-74.2}     & \textbf{-1.8}      & \textbf{-72.9}      & \textbf{-5.1}     & \textbf{-55.8}     \\ \hline
\end{tabular}
\label{tab: cifar data}
\end{table}

In Table \ref{tab: results}, \ref{tab: cifar data}, we present the significant outcomes of \our{}, providing insights into accuracy and cost metrics for images exhibiting varying levels of distortion. Notably, our proposed method, \our{}, demonstrates its remarkable performance by yielding the smallest performance drop ($\leq 5\%$) and the most substantial cost reduction ($> 65\%$) in comparison to the final exit. The cost values in the table are computed as the cumulative sum of both computational and offloading costs over the entire dataset.
It is worth noting that both the "Final Layer" and "Early Exit" baselines entail executing all computations on the mobile device, resulting in a significant computational cost that impacts the efficiency of the edge device. The accuracy loss observed in the "Early Exit" scenario primarily stems from misclassifications made by intermediate classifiers, which are subsequently rectified in the deeper layers. On the other hand, both "Random Exit" and "AdaEE" provide an option to offload, but they have fixed splitting layers, preventing dynamic adaptation of the splitting point. "AdaEE," however, possesses the advantage of adjusting thresholds based on contextual information.

Comparing our method with other baselines, \our{} showcases a remarkable reduction in cost by efficiently optimizing resource utilization on both edge and cloud platforms. The distinct advantage of \our{} lies in its adaptive nature, enabling it to select the optimal splitting layer based on past samples. It accounts for a minimal reduction in accuracy due to offloading to the cloud. Overall, \our{} effectively optimizes resource utilization across both mobile devices and cloud infrastructure by balancing the computational as well as communication costs.

\section{Conclusion}
We address the problem of executing the DNNs on edge devices by considering the resources available on the edge. We propose a method that combines early exits and split computing to reduce the computational cost of inference on edge devices as well as communication costs to the cloud. The developed algorithm \our{} splits the DNN based on the resources available where the initial part of the DNN is deployed on the edge device with an exit attached at the last layer on the edge device. If the sample gains sufficient confidence on the edge device, it is inferred locally, else it is offloaded to the cloud for inference on the final layer of DNN. \our{} adapts to changes in the distribution of the test data when it data arrives in an online and unsupervised. Notably, \our{} has a minimal drop in accuracy even when cost changes making it more robust to different user-specific devices.

Our work could be extended by simultaneously adapting to the optimal threshold as well as the splitting layer. Also, our method finds the optimal splitting layer based on all the samples, however, the splitting layer could be learned based on the difficulty level of the individual samples.

\bibliographystyle{IEEEtran}
\bibliography{mab,header}

\appendices

\end{document}